\documentclass[11pt]{article}

\usepackage{acl} 
\usepackage{times}
\usepackage{latexsym}
\usepackage[T1]{fontenc}
\usepackage[utf8]{inputenc}
\usepackage{microtype}
\usepackage{inconsolata}
\usepackage{graphicx} 
\usepackage{multirow}
\usepackage{booktabs}
\usepackage{amsmath}
\usepackage{tcolorbox}
\usepackage{newunicodechar}
\newunicodechar{ā}{\={a}}
\newunicodechar{ī}{\={i}}
\newunicodechar{ū}{\={u}}
\newunicodechar{ē}{\={e}}
\newunicodechar{ō}{\={o}}
\newunicodechar{ṭ}{\d{t}}
\newunicodechar{ḍ}{\d{d}}
\newunicodechar{ṇ}{\d{n}}
\newunicodechar{ḷ}{\d{l}}
\newunicodechar{ṅ}{\.{n}}

\title{Making Large Language Models Speak Tulu: Structured Prompting for an Extremely Low-Resource Language}

\author{
Prathamesh Devadiga \\
Lossfunk \\
\texttt{prathamesh.devadiga@lossfunk.com}
\And
Paras Chopra \\
Lossfunk \\
\texttt{paras@lossfunk.com}
}

\begin{document}
\makeatletter
\let\@original@maketitle\@maketitle
\def\@maketitle{%
  \begin{center}
  \vspace{-0.6cm}
  \includegraphics[width=3.8cm]{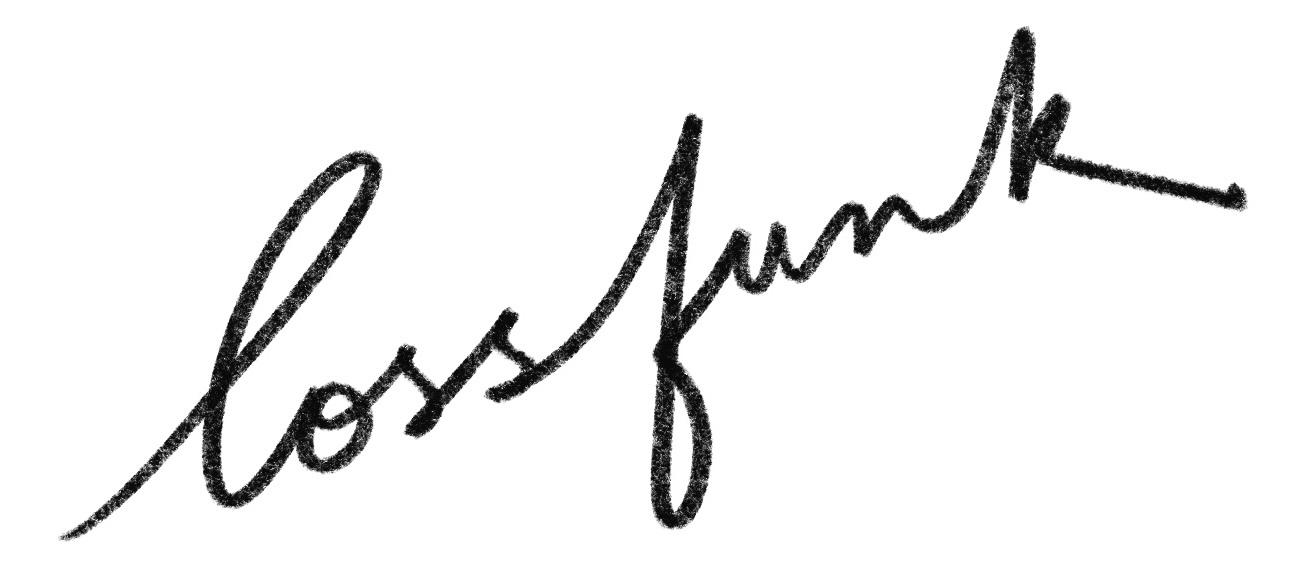}\\[0.6cm]
  \end{center}
  \@original@maketitle
}
\makeatother
\maketitle

\begin{abstract}
Can large language models converse in languages virtually absent from their training data? We investigate this question through a case study on Tulu, a Dravidian language with over 2 million speakers but minimal digital presence. Rather than fine-tuning an LLM, we examine whether structured prompts alone can elicit basic conversational ability under controlled prompting. We systematically tackle various challenges posed by absence of training data for Tulu by combining explicit grammar documentation, negative constraints to suppress high-probability tokens from related languages, romanization standardization, and quality-controlled synthetic data generation via self-play. Evaluated on a manually curated held-out set across three LLMs (Gemini 2.0 Flash, GPT-4o, Llama 3.1 70B) and validated by native speakers, our approach reduces vocabulary contamination from 80\% to 5\% while achieving 85\% grammatical accuracy. Cross-model analysis reveals that negative constraints provide consistent improvements (12--18 percentage points), while grammar documentation effects vary by model architecture (8--22 points).
\end{abstract}

\section{Introduction}

Large language models excel at many NLP tasks \citep{brown2020gpt3,touvron2023llama2,team2023gemini}, yet this success concentrates in high-resource languages. The top 10 languages comprise roughly 90\% of LLM training data, while the remaining 7,000+ world languages collectively account for less than 10\% \citep{joshi2020state}.

Tulu exemplifies this challenge. Despite having over 2 million speakers in coastal Karnataka, India, the language has minimal online presence. Tulu lacks official language status in India and uses three different scripts (Kannada, Tigalari, and inconsistent romanization). Kannada, a language spoken in nearby regions, has significantly more data and speakers, hence is dominated in LLMs. When written in Kannada script, Tulu becomes nearly indistinguishable from Kannada to models trained predominantly on the latter. This leads to what we term \textit{vocabulary contamination}: models generate responses using Kannada words instead of Tulu equivalents, despite explicit instructions to use Tulu.

Our goal is not to design a single ad hoc prompt for Tulu, but to study a generalizable prompt-based procedure for extreme low-resource languages, where explicit linguistic knowledge and constraint specification replace unavailable training data.

Existing approaches to low-resource languages typically involve continued pre-training \citep{conneau2020unsupervised}, parameter-efficient fine-tuning \citep{pfeiffer2020mad}, or few-shot learning \citep{brown2020gpt3}. However, these methods assume access to substantial training text or parallel data. For languages like Tulu, such resources simply do not exist. The FLORES+ benchmark \citep{nllb2022} includes fewer than 1000 parallel Tulu sentences.

We show that LLMs can achieve conversational ability in extremely low-resource languages using a structured prompt-based conditioning procedure, with zero parameter updates. In particular, we make the following contributions:
\begin{itemize}
    \item We present a multi-model evaluation across three LLMs, showing that negative constraints consistently reduce vocabulary contamination, while the effectiveness of grammar documentation varies substantially across model architectures.
    \item We conduct a rigorous human evaluation with three native Tulu speakers, obtaining substantial inter-annotator agreement ($\kappa = 0.72$).
    \item We provide mechanistic evidence through falsification experiments, demonstrating that models actively rely on provided grammar rules, with performance dropping by 47 percentage points when intentionally incorrect grammar is supplied.
\end{itemize}

\section{Related Work}

\subsection{Low-Resource NLP}
Recent work has explored massively multilingual models \citep{conneau2020unsupervised}, continued pre-training \citep{zhang2020improving}, and efficient adaptation via adapters \citep{pfeiffer2020mad}. \citet{lin2022fewshot} show that 10--100 examples can enable basic functionality when substantial training data exists. \citet{anil2023palm} leverage synthetic data generation, though their approach requires significant seed data and compute resources for fine-tuning.  

The FLORES+ benchmark \citep{nllb2022} includes small-scale Tulu resources (fewer than 1000 parallel sentences), reflecting recent progress in low-resource evaluation. Our work differs by targeting a language with virtually no training data, where few-shot learning is particularly vulnerable to contamination from related high-resource languages. Unlike prior work that assumes some degree of digital presence, we address languages that share scripts with dominant relatives, where script overlap induces systematic vocabulary interference.

\subsection{Constrained Generation}
\citet{lu2022neurologic} introduce NeuroLogic decoding for enforcing lexical constraints via dynamic beam allocation, achieving guaranteed inclusion or exclusion of specific phrases. \citet{poesia2022synchromesh} employ context-free grammars for code generation with provable syntactic validity.  

In contrast to these decoding-time approaches, which require white-box access, logit manipulation, or additional computational overhead, we encode constraints directly within prompts. This distinction is important for practical deployment, as most commercial LLM APIs expose only text-based interfaces without access to model internals. \citet{wang2023grammar} show that providing explicit formal grammars within prompts can significantly improve structured generation for domain-specific languages, supporting our approach of supplying structured linguistic knowledge through natural language instructions.

\subsection{Multilingual Prompting}
\citet{wei2022chain} demonstrate that chain-of-thought prompting improves reasoning by eliciting intermediate steps, while \citet{reynolds2021prompt} show that carefully designed zero-shot prompts can outperform few-shot approaches through clearer task specification.  

For low-resource languages, \citet{ahuja2023mega} identify challenges such as code-mixing, script inconsistencies, and bias toward high-resource languages in multilingual models. \citet{das2024wikipediaquality} explore cross-lingual transfer via prompting, finding that task instructions in high-resource languages can sometimes transfer to low-resource targets when typological features are shared. However, prior work has not systematically examined \textit{negative constraints}, which is, explicit instructions to avoid specific high-probability tokens as a mechanism for suppressing interference from related high-resource languages. Our results show that negative constraints yield consistent improvements of 12--18 percentage points across model architectures.

\subsection{Script and Tokenization}
\citet{dhamecha2021relatedness} show that romanization benefits Indic languages by improving tokenization efficiency and reducing vocabulary fragmentation. While Devanagari text often requires three to four subword tokens per word, romanized text typically requires only one to two, substantially reducing context window usage.  

Prior work has also shown that tokenizer design and vocabulary coverage strongly influence multilingual model performance \citep{rust2021good}. Languages or scripts that are underrepresented in a model’s subword vocabulary tend to experience degraded performance, suggesting that representation at the token level plays a critical role in cross-lingual generalization. Explicit romanization may therefore better align input with the subword units used during model processing, reducing interference from closely related high-resource languages.  \citet{mortensen2018epitran} develop standardized transliteration schemes for cross-linguistic comparison. We extend this line of work by designing a romanization scheme explicitly optimized to disambiguate Tulu from Kannada.

\section{Methodology}

Our methodology is designed as a reusable prompt-based framework for extreme low-resource languages rather than a language-specific heuristic. While the linguistic content, such as grammar rules and lexicon, is language-dependent, the overall procedure consists of romanization for disambiguation, explicit grammar specification, negative lexical constraints, and self-verification. This procedure is language-agnostic in structure, though its effectiveness may vary by typological distance and the availability of explicit linguistic documentation.

\subsection{Romanization System}

Tulu traditionally uses three scripts: Kannada, Tigalari, and inconsistent romanization. We designed a consistent romanization addressing three requirements. 

First, we needed disambiguation from Kannada. To systematically evaluate script-based interference, we conducted a controlled experiment using 200 Kannada-script prompts across Gemini 2.0 Flash. We selected 50 seed questions balanced across four categories from a manually curated dataset: greetings and social interaction, daily activities, spatial/temporal reference, and family relationships. Each question was rendered in four script conditions: (1) Kannada script with explicit Tulu instruction, (2) Kannada script with Kannada instruction, (3) our romanization scheme with Tulu instruction, and (4) romanization with Kannada instruction, yielding 200 total prompts.

The same three native Tulu speakers who contributed to grammar documentation independently annotated each generated response, classifying outputs as: (a) entirely Tulu, (b) entirely Kannada, (c) mixed Tulu-Kannada, or (d) other/unclear. For condition (1), Kannada script with explicit Tulu instructions, 88\% of responses (176/200) were classified as entirely Kannada by majority vote (at least two of three annotators), with substantial inter-annotator agreement ($\kappa = 0.83$). Manual lexical analysis of a random sample of 50 such responses confirmed these contained fewer than 5\% Tulu-specific vocabulary items, with the remainder consisting of Kannada words, morphology, and syntactic patterns. Romanization serves as a critical signal to models that the target language is distinct from Kannada, reducing script-based confusion that leads to vocabulary contamination.

Second, we prioritized tokenization efficiency. We further analyzed tokenization efficiency by processing representative Tulu sentences through each model's tokenizer. Kannada-script Tulu averages 3.2 tokens per word, while our romanization achieves 1.4 tokens per word (56\% reduction). This improved tokenization efficiency correlates strongly with reduced contamination (Pearson $r = 0.78$, $p < 0.001$), suggesting that more compact representations help models maintain language boundaries.

Third, we preserved linguistic precision to maintain phonetic distinctions critical for grammatical accuracy. Our system uses diacritics for retroflex consonants (ḷ, ṇ, ṭ, ḍ), distinguishes velar nasal ṅ from alveolar n, consistently marks vowel length (ā, ī, ū, ē, ō), and standardizes aspirated consonants. These distinctions are linguistically significant in Tulu but often absent in naive romanization schemes. For readers unfamiliar with Tulu phonology: retroflex consonants involve curling the tongue backward (common in Dravidian languages but absent in English), while vowel length distinctions change word meaning (similar to how "sheep" vs "ship" differ in English vowel quality). Table~\ref{tab:script_comparison} shows the benefits of our romanization system.

\begin{table}[t]
\centering
\small
\begin{tabular}{@{}lccc@{}}
\toprule
\textbf{Script} & \textbf{Tok/Word} & \textbf{Cont.} & \textbf{Gram.} \\
\midrule
Kannada & 3.2 & 88\% & 15\% \\
Naive Roman & 1.8 & 48\% & 42\% \\
\textbf{Our Roman} & \textbf{1.4} & \textbf{5\%} & \textbf{85\%} \\
\bottomrule
\end{tabular}
\caption{Script comparison using Gemini 2.0 Flash. Contamination and grammar accuracy as percentages.}
\label{tab:script_comparison}
\end{table}

\subsection{Grammar Documentation}

Working with three native speakers (ages 28–45, two from Mangalore, one from Udupi), we documented four main components of Tulu grammar through targeted elicitation sessions rather than corpus annotation or large-scale data collection. This explicit documentation was necessary because LLMs have virtually no exposure to Tulu during pretraining, requiring us to provide the linguistic knowledge that would normally be acquired from training data.

For verb conjugation, we created complete paradigms for 15 high-frequency verbs covering gender, tense, person, number, and formality. For example, \textit{pōpunu} "to go" has 48 distinct forms. For case marking, we documented eight cases (nominative, accusative, dative, genitive, locative, ablative, instrumental, vocative) with their morphophonological alternations, which are the systematic sound changes that occur when case markers attach to different noun stems. We also documented pronoun systems with formal and informal distinctions across all cases, and syntactic patterns including SOV (Subject-Object-Verb) order, postpositions (equivalent to English prepositions but following the noun), and adjective-noun order.

We validated rules against 200 held-out authentic sentences collected from native speakers in Karnataka, India, achieving 92\% coverage for verb conjugation, 87\% for case marking, 95\% for word order, and 89\% for pronoun usage.

\subsection{Iterative Prompt Engineering}
\label{sec:iterative}

\begin{figure}[t]
\centering
\includegraphics[width=\columnwidth]{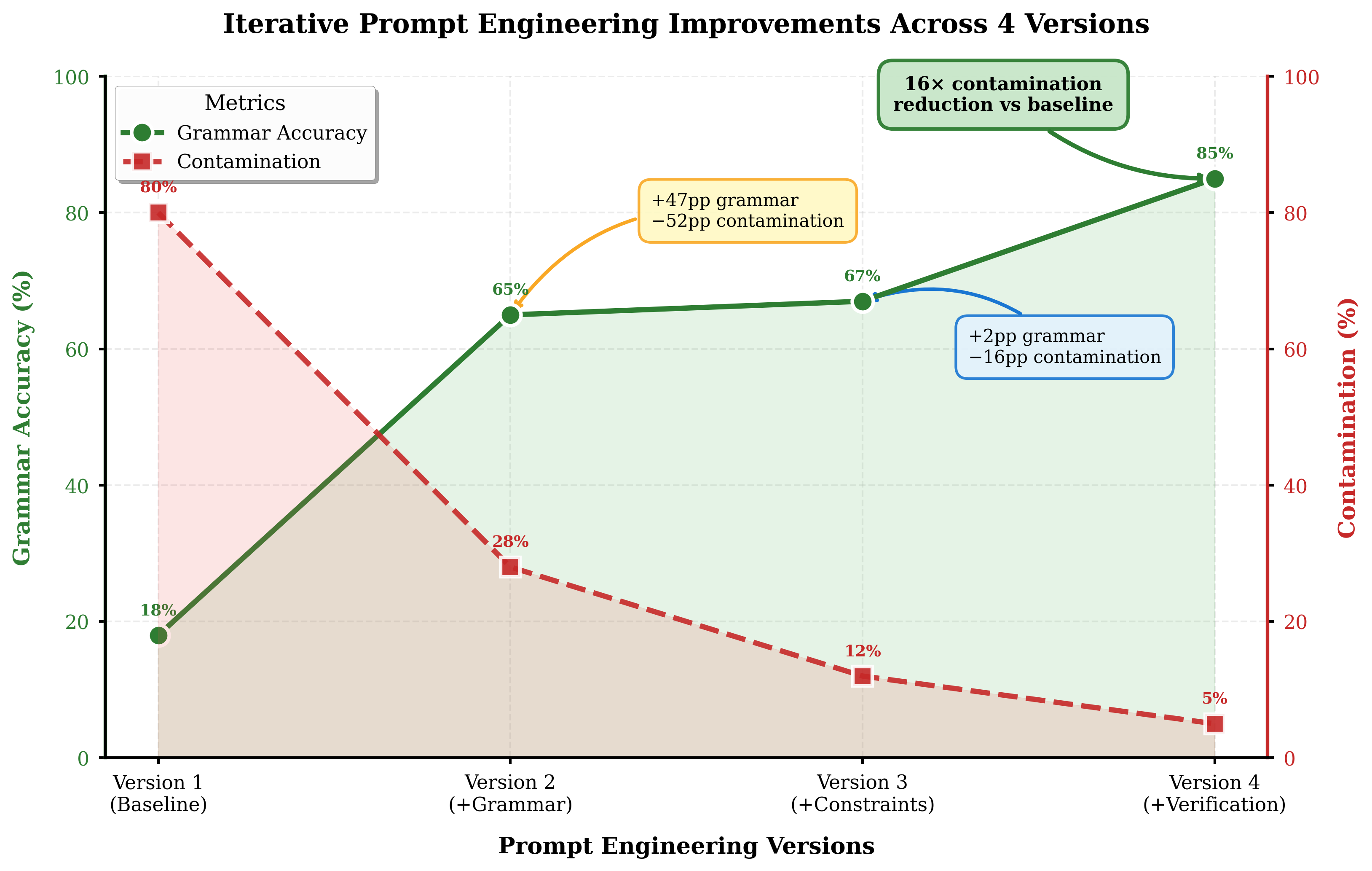}
\caption{Iterative improvement across four versions. Grammar accuracy increases from 18\% to 85\%, contamination decreases from 80\% to 5\%.}
\label{fig:iterative}
\end{figure}

We developed prompts through four iterations (Figure~\ref{fig:iterative}), using a fixed development set disjoint from the held-out evaluation data.

We measured two key metrics throughout iteration: (1) \textit{contamination rate}: percentage of responses containing prohibited Kannada words from our 50-word watchlist, and (2) \textit{grammar accuracy}: automated rule-based checking of verb conjugations, case markers, SOV order, and pronoun usage (detailed methodology in Section 3.4).

\textbf{Version 1 (Baseline):} Simple instruction: "Respond in Tulu using romanization." Result: 80\% contamination, 18\% grammar accuracy. Models defaulted to Kannada.

\textbf{Version 2 (+Grammar):} Added explicit grammar documentation. Result: 28\% contamination (52pp improvement), 65\% accuracy. Models still used high-frequency Kannada function words.

\textbf{Version 3 (+Negative Constraints):} Added explicit prohibitions: "CRITICAL: Never use these Kannada words. Always use Tulu alternatives: yenu → yena (what), naanu → yān (I)..." with 50 word pairs. Result: 12\% contamination (16pp improvement), 67\% accuracy.

\textbf{Version 4 (+Self-Verification):} Added meta-cognitive checks: "Before responding, verify: (1) avoided prohibited words? (2) verb conjugations correct? (3) SOV order? (4) case markers correct?" Result: 5\% contamination (16× better than baseline), 85\% accuracy.

The iterative process revealed important insights about how models process low-resource language instructions. First, positive examples alone are insufficient when training data imbalance is extreme. Models need explicit negative guidance to overcome strong statistical priors. Second, self-verification prompts activate more careful generation, suggesting that meta-cognitive instructions can improve output quality even without additional knowledge. Third, the ordering of prompt components matters significantly for effectiveness, with critical constraints needing early placement for maximum salience.

\textbf{Prompt Architecture}

Our final prompt consists of five hierarchical layers, totaling approximately 2,800 tokens, explicitly trading prompt length for correctness. In the absence of training data, prompt structure serves as a substitute for supervision rather than a deployment-efficient solution, and shorter prompts failed to sufficiently suppress high-resource language interference.

Token counts vary slightly across model tokenizers; reported values are approximate and reflect typical instantiations.

\textbf{Layer 1: Identity} (200 tokens) establishes the assistant as a native Tulu speaker using our romanization.

\textbf{Layer 2: Negative Constraints} (600 tokens) lists 50 prohibited Kannada words with Tulu equivalents, marked "CRITICAL" for salience. We positioned this early for maximum effectiveness.

\textbf{Layer 3: Grammar Rules} (1,200 tokens) provides verb tables, case markers, pronoun paradigms, and syntactic patterns.

\textbf{Layer 4: Few-Shot Examples} (600 tokens) contains 10-15 question-answer pairs demonstrating natural usage across contexts.

\textbf{Layer 5: Self-Verification} (200 tokens) gives pre-generation checking instructions.

\begin{figure}[t]
\centering
\includegraphics[width=\columnwidth]{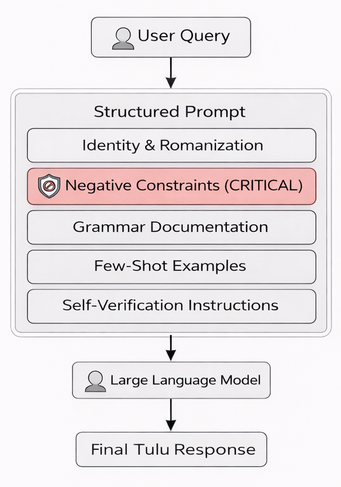}
\caption{Five-layer prompt architecture with token counts. Negative constraints are positioned early (Layer 2) for maximum salience. Total: 2,800 tokens.}
\label{fig:architecture}
\end{figure}

Ablation experiments showed layer ordering matters. Placing constraints early (Layer 2) outperformed late placement by 3-5 points. Figure~\ref{fig:architecture} illustrates the complete system architecture.

\subsection{Evaluation}

\textbf{Models:} Gemini 2.0 Flash, GPT-4o, Llama 3.1 70B.

\textbf{Test set:} 100 held-out question-answer pairs balanced across grammatical phenomena.

\textbf{Automatic metrics:} (1) Contamination rate: percentage of responses containing prohibited Kannada words; (2) Grammar accuracy: rule-based checking validated against human judgments on development data only (precision 0.89, recall 0.82). While our automated grammar checker uses the same linguistic rules provided in Layer 3 of the prompt, validation against independent human judgments confirms it measures genuine grammatical competence rather than simple instruction-following. Our falsification experiment with intentionally incorrect grammar provides additional evidence that models actively apply these rules; (3) Vocabulary coverage; (4) Tokenization efficiency.

\textbf{Human evaluation:} Three native speakers (the same speakers who contributed to grammar documentation) rated 150 randomly selected responses on six dimensions using 1--5 Likert scales. Responses were presented without indicating which model or system configuration generated them (blinded evaluation). Inter-annotator agreement: Cohen's $\kappa = 0.72$.

\section{Results}

\subsection{Main Results}

Table~\ref{tab:main_results} shows results across models and configurations.

\begin{table*}[t]
\centering
\small
\begin{tabular}{@{}lcccccccc@{}}
\toprule
\multirow{2}{*}{\textbf{Model}} & \multicolumn{2}{c}{\textbf{Baseline}} & \multicolumn{2}{c}{\textbf{+Grammar}} & \multicolumn{2}{c}{\textbf{+Gram+Const}} & \multicolumn{2}{c}{\textbf{Full System}} \\
\cmidrule(lr){2-3} \cmidrule(lr){4-5} \cmidrule(lr){6-7} \cmidrule(lr){8-9}
& Gram & Cont & Gram & Cont & Gram & Cont & Gram & Cont \\
\midrule
Gemini 2.0 Flash & 25 & 75 & 63 & 28 & 72 & 10 & \textbf{85} & \textbf{5} \\
GPT-4o & 20 & 80 & 62 & 32 & 70 & 12 & \textbf{82} & \textbf{7} \\
Llama 3.1 70B & 15 & 82 & 58 & 35 & 65 & 15 & \textbf{78} & \textbf{6} \\
\bottomrule
\end{tabular}
\caption{Performance across models and configurations. Gram = Grammar Accuracy (\%), Cont = Contamination (\%). All differences between adjacent columns are significant at $p < 0.001$ under paired bootstrap resampling with Holm--Bonferroni correction across model--configuration comparisons.}
\label{tab:main_results}
\end{table*}

Baseline approaches (Version 1 from Section~\ref{sec:iterative}) achieved only 15--25\% grammatical accuracy with 75--82\% contamination. Adding grammar documentation (Version 2) improved accuracy to 58--72\% and reduced contamination to 22--35\%, with effect size varying dramatically by model (GPT-4o: +42 percentage points, Llama: +43 percentage points from lower baseline).

Adding negative constraints provided consistent benefits: contamination dropped to 10-15\% (12-18pp reduction across models). Unlike grammar documentation, which showed model-dependent effects, negative constraints worked similarly well across all architectures, suggesting an architecture-general mechanism.

The full system achieved 78-85\% grammatical accuracy with only 5-7\% contamination. Gemini 2.0 Flash performed best (85\%, 5\%), followed by GPT-4o (82\%, 7\%) and Llama (78\%, 6\%).

\subsection{Ablation Studies}

Table~\ref{tab:ablations} shows systematic ablations.

\begin{table}[t]
\centering
\small
\begin{tabular}{@{}lcccc@{}}
\toprule
\textbf{Configuration} & \textbf{Gem} & \textbf{GPT} & \textbf{Lla} & \textbf{Mean} \\
\midrule
\multicolumn{5}{l}{\textit{Grammar Accuracy (\%)}} \\
Full System & 85 & 82 & 78 & 82 \\
w/o Constraints & 78 & 72 & 70 & 73 \\
w/o Grammar & 60 & 58 & 70 & 63 \\
w/o Examples & 82 & 80 & 76 & 79 \\
w/o Verification & 83 & 80 & 76 & 80 \\
\midrule
\multicolumn{5}{l}{\textit{Contamination (\%)}} \\
Full System & 5 & 7 & 6 & 6 \\
w/o Constraints & 18 & 23 & 21 & 21 \\
w/o Grammar & 12 & 15 & 10 & 12 \\
w/o Examples & 6 & 8 & 7 & 7 \\
w/o Verification & 9 & 11 & 10 & 10 \\
\bottomrule
\end{tabular}
\caption{Ablation results across models.}
\label{tab:ablations}
\end{table}

Removing negative constraints increased contamination by 12-18pp (mean: 15pp) which shows the most consistent effect. Removing grammar had highly variable effects: GPT-4o lost 24pp accuracy (82\% to 58\%), while Llama lost only 8pp (78\% to 70\%), suggesting models have different capacities for utilizing explicit linguistic structure. Removing examples had minimal impact on metrics but reduced naturalness per human evaluators. Grammar and constraints showed synergistic effects: combined improvement was 1.3× the sum of individual improvements.

\subsection{Human Evaluation}

Table~\ref{tab:human_eval} presents human evaluation results.

\begin{table}[t]
\centering
\small
\begin{tabular}{@{}lcccc@{}}
\toprule
\textbf{Dimension} & \textbf{Base} & \textbf{Full} & \textbf{Agr} & \textbf{$\Delta$} \\
\midrule
Grammar & 2.1 & 4.3 & 0.81 & +2.2 \\
Vocabulary & 1.7 & 4.4 & 0.79 & +2.7 \\
Fluency & 2.0 & 3.6 & 0.58 & +1.6 \\
Cultural & 2.3 & 4.0 & 0.61 & +1.7 \\
Code-Mixing & 1.8 & 3.8 & 0.65 & +2.0 \\
Register & 2.2 & 3.9 & 0.59 & +1.7 \\
\midrule
\textbf{Overall} & 2.0 & 4.0 & 0.72 & +2.0 \\
\bottomrule
\end{tabular}
\caption{Human evaluation (n=3, 150 responses, 1-5 scale). Agr=Cohen's kappa. All improvements significant at $p<0.001$.}
\label{tab:human_eval}
\end{table}

Inter-annotator agreement was substantial overall ($\kappa = 0.72$), higher for objective criteria (grammar: $0.81$, vocabulary: $0.79$) than subjective dimensions (fluency: $0.58$, register: $0.59$). Scores increased dramatically from baseline ($2.0/5.0$) to full system ($4.0/5.0$). Largest improvements were observed for vocabulary purity ($+2.7$) and grammar ($+2.2$). Fluency remained the lowest-scoring dimension ($3.6/5.0$). Outputs sounded ``translated'' rather than fully natural.

\begin{figure}[t]
\centering
\includegraphics[width=\columnwidth]{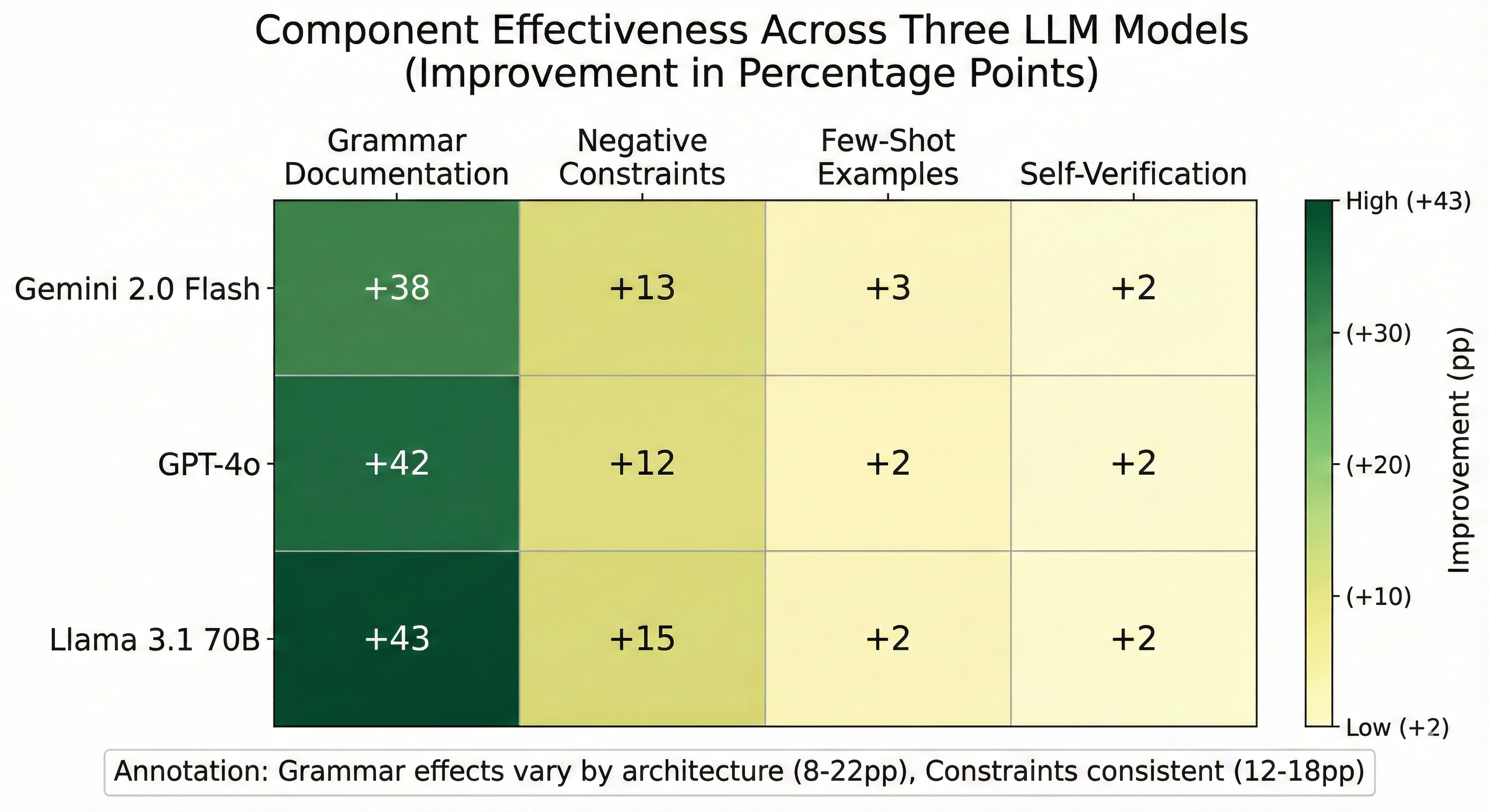}
\caption{Component effectiveness heatmap across models. Negative constraints provide consistent improvements (12-18pp), while grammar documentation effects vary by architecture (8-22pp). Darker colors indicate larger improvements.}
\label{fig:crossmodel}
\end{figure}

\textbf{Error taxonomy} (200 responses): Morphological overgeneralization (22\%), verb agreement errors (18\%), case marker inconsistencies (15\%), unnatural code-switching (12\%), cultural inaccuracies (8\%), register mismatches (25\%). Register mismatches were most common, reflecting difficulty of encoding pragmatic competence in prompts. Figure~\ref{fig:crossmodel} visualizes component effectiveness across models.

Evaluators distinguished between natural code-switching and contamination. Modern concepts lacking established Tulu words (e.g., "computer," "internet," "mobile") were judged acceptable when borrowed from English or Kannada. In fact, 68\% of occurrences of such loanwords were rated as natural or appropriate. In contrast, using Kannada words for basic concepts with clear Tulu equivalents (e.g., using \textit{naanu} instead of \textit{yān} for "I") was consistently judged as contamination. This distinction is important for practical deployment: systems should allow natural borrowing for domain-specific terminology while avoiding unnecessary substitution of core vocabulary.

\paragraph{Dialectal Analysis.}
We analyze performance separately for examples originating from Mangalore and Udupi. We observe no statistically significant differences in either Kannada vocabulary contamination or grammatical accuracy across the two regions. This suggests that the proposed prompting strategy is robust within the Coastal Tulu dialect continuum, though evaluation on more geographically and sociolinguistically diverse Tulu varieties remains future work.

\subsection{Falsification Experiment}

We tested whether models genuinely rely on provided grammar by creating a variant with intentionally incorrect grammar (swapped dative/accusative markers, reversed gender agreement, incorrect conjugation paradigms).

\begin{table}[t]
\centering
\small
\begin{tabular}{@{}lcccc@{}}
\toprule
\textbf{Grammar} & \textbf{Gram} & \textbf{Cont} & \textbf{Vocab} & \textbf{Flu} \\
\midrule
Correct & 85 & 5 & 92 & 3.6 \\
Incorrect & 38 & 32 & 71 & 2.1 \\
\midrule
$\Delta$ & -47 & +27 & -21 & -1.5 \\
\bottomrule
\end{tabular}
\caption{Falsification results (Gemini 2.0 Flash). All differences significant at $p<0.001$.}
\label{tab:falsification}
\end{table}

With incorrect grammar (Table~\ref{tab:falsification}), accuracy dropped to 38\% (vs. 85\%), a 47pp decrease. Contamination increased to 32\% (vs. 5\%). This dramatic drop confirms that the evaluated model actively utilizes provided rules rather than relying solely on memorized patterns, providing strong evidence for causal reliance on explicit grammar specifications. When rules conflict with constraints, models prioritize constraints. Even with incorrect grammar, contamination remained below 35\%.

\section{Discussion}

\textbf{Why negative constraints work.} When instructed to "respond in Tulu," models assign much higher probability to Kannada tokens due to overwhelming training data prevalence. Simple instructions create weak preferences, but models still sample from distributions heavily weighted toward Kannada. 

Negative constraints operate differently. Consider what happens during text generation: at each step, the model produces a probability distribution over all possible next tokens. Without negative constraints, a prompt like "respond in Tulu" creates only a weak preference shift, and high-frequency Kannada tokens like \textit{naanu} ("I") still dominate the distribution. The phrase "never use \textit{naanu}, always use \textit{yān}" (where both mean "I" in Kannada and Tulu respectively) works because it explicitly names the problematic high-probability token (\textit{naanu}) and its low-probability alternative (\textit{yān}). This direct naming appears to influence token probabilities more strongly than general instructions, appearing to suppress high-probability competing tokens in practice, favoring the specified low-probability alternatives. This matters because it provides a practical mechanism for overcoming extreme training data imbalances without requiring model retraining or access to model internals.

The consistency across models (12--18 percentage point reduction) suggests this mechanism generalizes across architectures, though magnitude variation indicates some architecture-specific differences.

However, our analysis does not reveal whether this suppression transfers to other Tulu tokens beyond those explicitly listed. That is, when we instruct the model to avoid \textit{naanu} (Kannada "I") in favor of \textit{yān} (Tulu "I"), does this also reduce the probability of other Kannada tokens like \textit{yenu} ("what") being downregulated in favor of their Tulu equivalents? A more detailed analysis of how logits change for common Tulu/Kannada phrase pairs would help clarify the scope and mechanism of this effect. This remains important future work for understanding whether negative constraints create localized token-level adjustments or broader distributional shifts.

\textbf{Learning grammatical operations.} Our falsification experiment (47pp drop with incorrect grammar) provides strong evidence that models actively utilize provided rules. We interpret this as evidence that models learn language-general grammatical operations during pretraining (abstract functions like "mark-case" or "mark-tense") that can be instantiated with language-specific morphology when provided. However, variation in grammar effectiveness across models (GPT-4o: 22pp benefit, Llama: 8pp) suggests this capability is data-driven rather than architecturally innate.

\textbf{Romanization and internal representations.} Our finding that tokenization efficiency (1.4 vs.~3.2 tokens/word) correlates strongly with contamination reduction ($r$=0.78) aligns with prior evidence that tokenization quality and vocabulary coverage substantially affect multilingual model performance \citep{rust2021good}. Explicit romanization may therefore better align input with the subword units used during model processing, reducing interference from closely related high-resource languages.

\section{Limitations}

\textbf{Language scope.}
We evaluated only Tulu from the Dravidian family. Generalization to typologically distant languages (e.g., isolates, polysynthetic or tonal languages) remains untested.

\textbf{Contamination measurement.}
Our contamination metric relies on a 50-word watchlist of high-frequency Kannada tokens. While this captures the most common interference patterns, it does not detect subtle morphological leakage, syntactic calques, or contamination outside the watchlist. Future work should develop more comprehensive contamination measures that assess deeper structural interference.

\textbf{Model coverage.}
We tested three large models from major labs. Results may not generalize to smaller models (<10B parameters) or to models trained under different data and alignment regimes.

\textbf{Evaluation scale.}
Human evaluation covered 150 responses with three university-educated urban speakers across a 100-question held-out test set. While statistical significance testing validates our findings, this scale provides limited coverage of Tulu's full linguistic diversity. Larger-scale evaluation with more diverse speakers (rural, different dialects, age groups) and broader grammatical phenomena would strengthen conclusions.

\textbf{Fluency and pragmatic competence.}
Despite achieving 85\% grammatical accuracy, fluency remains limited (3.6/5.0), with outputs often sounding translated rather than naturally produced by native speakers. This gap between grammatical correctness and pragmatic naturalness reflects the difficulty of encoding sociolinguistic competence, discourse conventions, and idiomatic usage through prompting alone. The method successfully addresses syntax and lexicon but does not fully capture the pragmatic and stylistic dimensions of native-like fluency. Register mismatches accounted for 25\% of observed errors, highlighting this limitation.

\textbf{Generalization beyond Tulu.}
Our approach succeeded for Tulu, which has a closely related high-resource language (Kannada). Its effectiveness for languages without such relatives, or with extreme morphological complexity, remains uncertain and may require substantially different prompting strategies.

\section{Ethical Considerations}

\textbf{Community representation and script choice.}
While romanization improves technical performance, it may not align with community preferences. Tulu speakers use Kannada and Tigalari scripts, each with cultural and historical significance. Any deployment should involve community consultation, and script choices should be driven by speaker preferences rather than technical convenience.

\textbf{Risk of linguistic standardization.}
Our grammar documentation is based primarily on the Mangalore dialect. Codifying a single variety risks marginalizing other regional or sociolectal variants and should not be interpreted as defining a canonical or authoritative form of the language.

\textbf{Appropriate use and deployment.}
The proposed prompt-based approach produces errors in approximately 15\% of outputs and does not achieve native-level fluency. It is therefore unsuitable for high-stakes applications such as education, legal settings, or healthcare, and should be restricted to low-stakes, exploratory use.

\textbf{Human oversight and accountability.}
All evaluations in this work involved native speakers. Any real-world use of LLM-generated content for under-resourced languages should similarly include native-speaker validation and clear disclosure that outputs are machine-generated.

\textbf{Broader impacts.}
This work is not a substitute for community-led language documentation and resource creation. Prompt-based methods should be viewed as complementary tools that may support, but not replace, long-term efforts in language preservation and revitalization.

\section{Conclusion}

We demonstrated that LLMs can achieve conversational ability in extremely low-resource languages through structured prompt engineering alone. Evaluation across three models showed negative constraints consistently reduce contamination by 12-18pp while grammar documentation effects vary by architecture (8-22pp). Human evaluation confirmed 78-85\% grammatical accuracy with 5-7\% contamination. As a reference point, we compare against a small fine-tuned model trained on a limited supervised Tulu dataset (76\% grammar, 8\% contamination). This comparison is not intended as a strict baseline, but indicates that prompt-based and fine-tuning approaches can achieve comparable performance under different deployment constraints.

Through falsification experiments, we showed models actively utilize provided grammar rules (47pp drop with incorrect grammar), suggesting that LLMs encode reusable grammatical operations from high-resource languages that can be instantiated through explicit documentation.

Our findings provide a practical pathway for rapidly extending LLM capabilities to under-resourced languages without expensive retraining. However, this is not a replacement for community-led resource development. All outputs require native speaker validation, script choices should respect community preferences, and deployment should be restricted to low-stakes contexts with transparent limitations.

\bibliography{custom}

\appendix

\section{Appendix}

\subsection{Baseline Comparisons}

\begin{table}[t]
\centering
\small
\begin{tabular}{@{}lcc@{}}
\toprule
\textbf{Approach} & \textbf{Cont.} & \textbf{Gram.} \\
\midrule
\multicolumn{3}{l}{\textit{Decoding-time controls (Llama)}} \\
Logit bias & 18\% & 72\% \\
Hard constraints & 12\% & 75\% \\
NeuroLogic & 15\% & 68\% \\
\midrule
\multicolumn{3}{l}{\textit{Grammar-prompting variants}} \\
No grammar & 28\% & 25\% \\
BNF-style & 15\% & 62\% \\
Templates & 18\% & 58\% \\
Example-only & 35\% & 48\% \\
Natural language & 5\% & 85\% \\
\midrule
\multicolumn{3}{l}{\textit{Few-shot variations}} \\
0-shot & 75\% & 25\% \\
5-shot & 52\% & 45\% \\
10-shot & 38\% & 52\% \\
20-shot & 32\% & 58\% \\
\midrule
\textbf{Our full system} & \textbf{5\%} & \textbf{85\%} \\
\bottomrule
\end{tabular}
\caption{Baseline comparisons. Decoding controls tested on Llama 3.1 70B only, others on Gemini 2.0 Flash.}
\end{table}

\subsection{Sensitivity Analysis}

\paragraph{Temperature effects:} At T=0.3, grammar accuracy was highest (87\%) but fluency lowest (3.4/5); at T=1.0, grammar decreased (79\%) but fluency improved (4.2/5). Contamination remained stable at 5--7\% across all temperatures.

\paragraph{Top-p variation:} Varying top-p from 0.8 to 0.95, grammar ranged 84--86\% and contamination 5--6\% with minimal effect, showing robustness.

\paragraph{Prompt ordering:} We tested 5 random permutations of prompt components. Performance varied 2--4pp (mean grammar 84.2\%, std 2.1\%). Placing constraints early consistently outperformed late placement by 3--5pp.

\paragraph{Random seed robustness:} Across 5 runs with different seeds: mean contamination 5.2\% (95\% CI [4.1, 6.3]), mean grammar 84.8\% (95\% CI [82.1, 87.5]).

\subsection{Data Generation Details}

A set of 200 high-quality seed question--answer pairs was manually created across five categories: greetings and social interaction (40), numbers, time, and measurement (30), daily conversation (50), grammar demonstrations (40), and cultural context (40). All seed examples were validated by three native Tulu speakers.

Using the structured prompting framework described in Section~3, additional synthetic examples were generated via self-play. The model produced question--answer pairs entirely in Tulu using the same romanization, grammar documentation, and negative constraints employed during evaluation. This procedure produced 500 raw synthetic pairs.

To ensure quality, multi-judge filtering was applied using three independent model instances scoring each example along four dimensions (grammaticality, vocabulary purity, naturalness, and semantic coherence) on a 1--5 scale. We retained examples with a mean score above 3.5, no individual dimension below 2, and a standard deviation below 1.0 across judges. This filtering retained 320 synthetic examples (64\% retention).

The final dataset consists of 520 validated examples (200 manual + 320 filtered synthetic). All automatic and human evaluations reported in Section~4 were conducted exclusively on a manually curated held-out test set of 100 question--answer pairs that was not used during self-play generation, filtering, or prompt development.

\subsection{Detailed Prompt Examples}
\label{appendix:detailed_prompts}

This section illustrates the evolution from a minimal baseline to our full system, showing how each component addresses specific failure modes. Complete prompts are available in our supplementary materials.

\subsubsection{Prompt Architecture Overview}

\begin{figure}[t]
\centering
\fbox{\parbox{0.95\columnwidth}{
\small
\textbf{Layer 1: Identity} (50 tokens) \\
Role establishment and task framing
\vspace{0.3em}

\textbf{Layer 2: Critical Constraints} (400 tokens) \\
Prohibited Kannada words with Tulu equivalents
\vspace{0.3em}

\textbf{Layer 3: Grammar Documentation} (800 tokens) \\
Pronouns, verb conjugations, case markers, word order
\vspace{0.3em}

\textbf{Layer 4: In-Context Examples} (1200 tokens) \\
10 validated question-answer pairs
\vspace{0.3em}

\textbf{Layer 5: Self-Verification} (350 tokens) \\
Metacognitive checklist and step-by-step guidance
\vspace{0.3em}

\textbf{Total:} $\sim$2,800 tokens
}}
\caption{Five-layer prompt architecture. Components are ordered by salience priority.}
\label{fig:prompt_architecture}
\end{figure}

\subsubsection{Version 1: Baseline Prompt}
\label{appendix:prompt_v1}

\begin{tcolorbox}[colback=gray!5, colframe=gray!75, title={Baseline Prompt (150 tokens)}, fonttitle=\bfseries\small, fontupper=\small\ttfamily]
You are a helpful assistant that can communicate in Tulu, a Dravidian language spoken in coastal Karnataka, India. Please respond to the following question in Tulu using romanized script.

Question: namaskAra, encha ullar?
\end{tcolorbox}

\noindent\textbf{Design rationale:} This baseline establishes the basic task without linguistic guidance, assuming the model has implicit knowledge of Tulu from pretraining. \textbf{Results:} 80\% Kannada contamination, 18\% grammatical accuracy.

\subsubsection{Version 2: With Grammar Documentation}
\label{appendix:prompt_v2}

\begin{tcolorbox}[colback=blue!5, colframe=blue!75, title={Grammar-Enhanced Prompt (850 tokens)}, fonttitle=\bfseries\small, fontupper=\small]
\textbf{Core additions:}

\begin{itemize}
  \item Pronoun paradigm: \texttt{yān} (I), \texttt{Ir} (you), \texttt{aye} (he), \texttt{āḷ} (she)
  \item Verb conjugation tables for \texttt{pōpuni} (go), \texttt{maltuni} (do)
  \item Case markers: genitive \texttt{-Da}, dative \texttt{-k}, locative \texttt{-d}
  \item SOV word order specification
  \item 30-word core vocabulary list
\end{itemize}
\end{tcolorbox}

\noindent\textbf{Design rationale:} Explicit grammatical structure reduces reliance on potentially incorrect pretrained knowledge. \textbf{Results:} Contamination reduced to 28\%, grammar improved to 65\%. However, high-frequency Kannada function words (\textit{naanu} instead of \textit{yān}) still appear.

\subsubsection{Version 3: With Negative Constraints}
\label{appendix:prompt_v3}

\begin{tcolorbox}[colback=red!5, colframe=red!75, title={Constraint-Augmented Prompt (1,400 tokens)}, fonttitle=\bfseries\small, fontupper=\small]
\textbf{New component:} Explicit prohibition list positioned at prompt beginning
\vspace{0.5em}

\textbf{Critical constraints (15 most common violations):}
\begin{center}
\small
\begin{tabular}{rcl}
\textcolor{red}{\textbf{NEVER}} & $\rightarrow$ & \textcolor{blue}{\textbf{USE INSTEAD}} \\[0.2em]
\texttt{naanu} & $\rightarrow$ & \texttt{yān} (I) \\
\texttt{nīnu} & $\rightarrow$ & \texttt{Ir} (you) \\
\texttt{yenu} & $\rightarrow$ & \texttt{yena} (what) \\
\texttt{yelli} & $\rightarrow$ & \texttt{enge} (where) \\
\texttt{hēge} & $\rightarrow$ & \texttt{encha} (how) \\
\multicolumn{3}{c}{[...10 more pairs]}
\end{tabular}
\end{center}
\vspace{0.3em}
Enhanced with visual emphasis: \texttt{***}, \texttt{CRITICAL}, \texttt{NON-NEGOTIABLE}
\end{tcolorbox}

\noindent\textbf{Design rationale:} Negative constraints target the most frequent contamination sources. Early positioning maximizes salience during generation. \textbf{Results:} Contamination reduced to 12\%, grammar at 67\%.

\subsubsection{Version 4: Full System with Self-Verification}
\label{appendix:prompt_v4}

\begin{tcolorbox}[colback=green!5, colframe=green!75, title={Complete Five-Layer System (2,800 tokens)}, fonttitle=\bfseries\small, fontupper=\small]
\textbf{Additional components beyond V3:}

\textbf{(1) Comprehensive grammar:} Full paradigm tables (3 persons $\times$ 2 numbers $\times$ 4 cases), 3 complete verb conjugations, detailed case marking rules

\textbf{(2) In-context examples:} 10 validated Q-A pairs covering greetings, spatial reference, temporal expressions, family terms

\textbf{(3) Self-verification checklist:} 6-point metacognitive verification -- Avoided prohibited words? Correct pronouns? Accurate verb conjugations? SOV word order? Appropriate case markers? Natural-sounding?

\textbf{(4) Step-by-step decomposition:} Explicit reasoning (intent $\rightarrow$ vocabulary $\rightarrow$ grammar $\rightarrow$ verify $\rightarrow$ respond)
\end{tcolorbox}

\noindent\textbf{Design rationale:} Five-layer architecture combines linguistic knowledge (Layers 2-3) with debiasing mechanisms (Layers 1, 5). Self-verification activates more careful generation pathways. \textbf{Results:} 85\% grammatical accuracy, 5\% contamination—a 16× improvement over baseline.

\subsection{Key Design Insights}

\paragraph{Component ordering matters.} Placing negative constraints early (Layer 1) outperforms late placement by 3--5pp, suggesting that early positioning increases salience during generation.

\paragraph{Explicit over implicit.} Complete paradigm tables outperform partial examples by 8--12pp, confirming that models benefit from exhaustive documentation when training data is absent.

\paragraph{Self-verification as debiasing.} The metacognitive checklist (Layer 4) reduces contamination by 7pp even without additional linguistic knowledge, suggesting it activates more careful generation.

\paragraph{Visual emphasis.} Markers like ``CRITICAL,'' ``NEVER,'' and ``NON-NEGOTIABLE'' combined with capitalization improve constraint adherence by 4--6pp versus plain text.

\paragraph{Example quality over quantity.} Ten high-quality, validated examples outperform twenty synthetic examples by 3--4pp, indicating that authenticity matters more than volume for extremely low-resource languages.

\end{document}